\pdfoutput=1

\documentclass[letterpaper, 10 pt, conference]{ieeeconf}

\IEEEoverridecommandlockouts
\usepackage{amsmath}
\usepackage{amssymb}

\usepackage{tabularray}

\usepackage{algorithm}
\usepackage{algpseudocode}
\usepackage{subcaption}
\usepackage{graphicx}
\usepackage{balance}

\usepackage{booktabs}
\usepackage{multirow}
\usepackage{colortbl}
\usepackage{siunitx}
\usepackage[dvipsnames]{xcolor}

\makeatletter
\let\NAT@parse\undefined
\makeatother

\makeatletter
\newcommand*\titleheader[1]{\gdef\@titleheader{#1}}
\AtBeginDocument{%
	\let\st@red@title\@title
	\def\@title{%
		\bgroup\normalfont\large\centering\@titleheader\par\egroup
		\vskip1.5em\st@red@title}
}
\makeatother

\usepackage[hidelinks]{hyperref}

\titleheader{This work has been submitted to the IEEE for possible publication. Copyright may be transferred without notice, after which this version may no longer be accessible.}

\title{\LARGE \bf
Fast Coordinated Bimanual Motion Planning With Hard Constraints
}

\author{Borna Paro, Luka Petrović, and Ivan Marković
\thanks{All authors are with the University of Zagreb Faculty of Electrical Engineering and Computing, Laboratory for Autonomous Systems and Mobile Robotics (LAMOR), Unska 3, HR-10000, Zagreb, Croatia
        {\tt\small \{name.surname\}@fer.unizg.hr}}%
}

\begin{document}

\maketitle
\thispagestyle{empty}
\pagestyle{empty}

\setlength{\abovedisplayskip}{6pt plus 2pt minus 2pt}
\setlength{\belowdisplayskip}{6pt plus 2pt minus 2pt}
\setlength{\abovedisplayshortskip}{3pt plus 1pt minus 1pt}
\setlength{\belowdisplayshortskip}{3pt plus 1pt minus 1pt}
\setlength{\textfloatsep}{8pt plus 2pt minus 2pt}
\setlength{\floatsep}{8pt plus 2pt minus 2pt}
\setlength{\intextsep}{8pt plus 2pt minus 2pt}

\begin{abstract}
Bimanual manipulation enables complex tasks but introduces added complexity from the high number of degrees of freedom involved. When handling rigid objects, the relative transformation between the two end effectors must remain fixed throughout the motion, manifesting as a nonlinear equality constraint that confines the feasible configuration space to a measure-zero manifold and challenges conventional motion planners. We propose a fast bimanual motion planning pipeline that enforces this hard transformation constraint continuously along the entire path, using a leader-follower parameterization: the leader's configuration is treated as a free variable, while the follower's is determined via inverse kinematics to satisfy the constraint. We extensively evaluate the method in simulation across diverse environments, constraints and bimanual platforms, achieving 19.4× faster planning than prior work while guaranteeing continuous constraint satisfaction. Real-world experiments on a bimanual Kinova Gen3 setup, involving tray transport and elongated-object manipulation, validate direct transfer of planned trajectories to physical hardware. Project page with supplementary videos: \url{https://bornaparo.github.io/coordinated-bimanual-motion-planning/}\footnote{An open-source implementation of the proposed method will be made publicly available upon completion of the review process.}.

\end{abstract}

\section{INTRODUCTION}
Bimanual manipulation is a fundamental capability for robots operating in human-centric environments, requiring explicit spatial and temporal coordination between both arms~\cite{bimanual-manipulation-taxonomy}.
A wide range of practically relevant tasks can only be accomplished when two arms work in concert: folding laundry, assembling furniture, or moving a large panel through a doorway~\cite{bimanual-manipulation-taxonomy, holladay2024}.
Industrial applications likewise increasingly demand bimanual coordination for handling heavy, oversized, or irregularly shaped parts that exceed the reach or payload capacity of a single manipulator.

Bimanual tasks exhibit rich diversity in how the two arms interact. In many scenarios, the arms assume asymmetric roles, with one arm stabilizing a workpiece while the other performs a fine manipulation action; in other scenarios, both arms coordinate symmetrically to achieve a shared objective, for example when two hands cooperatively lift and transport a large box~\cite{bimanual-manipulation-taxonomy, holladay2024}. Symmetric cooperative transport, in particular, gives rise to rigid kinematic coupling between the end effectors: when both arms rigidly grasp a common object, the transformation between the two end effectors must remain fixed during transport. This requirement manifests as a nonlinear equality constraint in the configuration space (C-space) of the bimanual system, which is particularly challenging for motion planning algorithms. The set of configurations satisfying the constraint forms a measure-zero subset of the full C-space, so randomly sampled configurations will almost surely violate the constraint~\cite{sampling-based-methods-constraints-survey}, while trajectory optimization methods may struggle to converge to a feasible solution due to the nonconvexity of the constraint~\cite{traj-opt-manifold, sampling-based-methods-constraints-survey}.

\begin{figure}[!t]
    \centering
    \begin{subfigure}[t]{0.43\linewidth}
        \centering
        \includegraphics[width=\linewidth]{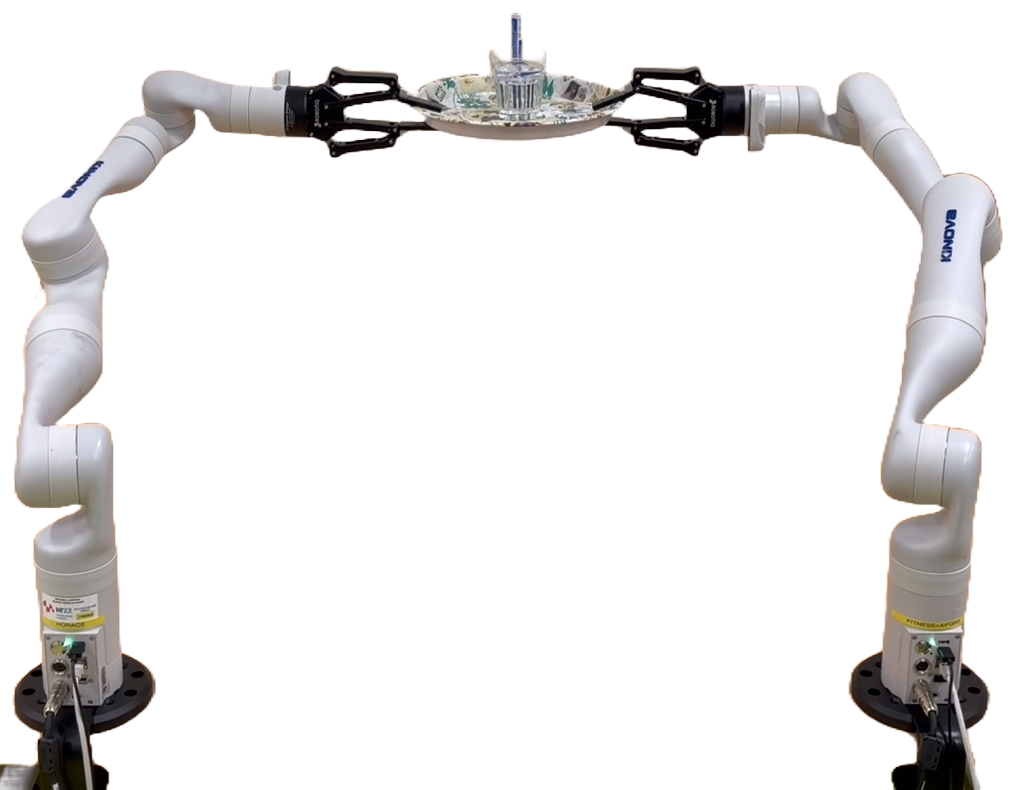}
        \caption{}
        \label{fig:tacna-exp-setup}
    \end{subfigure}
    \hfill
    \begin{subfigure}[t]{0.49\linewidth}
        \centering
        \includegraphics[width=\linewidth]{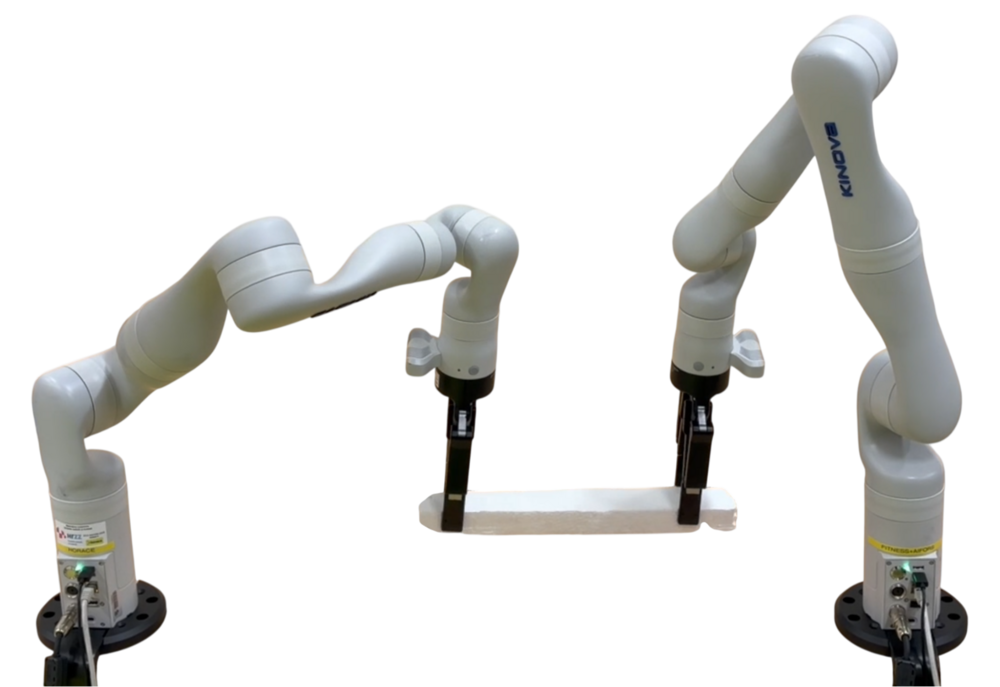}
        \caption{}
        \label{fig:tacna-exp-motion-trail}
    \end{subfigure}

    \caption{Real-world demonstrations of constrained coordinated bimanual motion planning. The two robot arms jointly transport (a) a tray while maintaining it upright to prevent spilling the water and tipping the object, and (b) an elongated object while preserving the coordinated dual-arm grasp.}
    \label{fig:tacna-exp-first}
\end{figure}

A key practical motivation for constrained bimanual planning arises in cooperative transport tasks where maintaining the relative end-effector pose is physically necessary. Consider transporting a tray loaded with a glass of water and upright objects as shown in Fig. \ref{fig:tacna-exp-first}; even a small deviation in the relative transformation between the grippers could tilt the tray, causing the water to spill or objects to topple. Similarly, when two arms carry an elongated object that is too long or heavy for a single arm, the rigid coupling must be maintained precisely to prevent the object from slipping or deforming. 

In this work, we present a fast bimanual motion planning pipeline that enforces a hard rigid-body transformation constraint between the two end effectors at every point along the planned trajectory.
The proposed method adopts a leader-follower parameterization in which the leader's configuration is treated as the free variable, while the follower's configuration is determined via inverse kinematics (IK) to satisfy the constraints.
In contrast to prior work~\cite{tedrake-constrained-bimanual}, the proposed method does not require offline precomputation or analytic solution to the IK problem, making it applicable to arbitrary manipulators and deployable immediately in unseen environments.
Our main contributions are as follows:
\begin{enumerate}
    \item A robot-agnostic motion planning pipeline that produces a smooth, executable trajectory in the bimanual C-space, while requiring as inputs only the start and goal poses and the desired transformation constraint.
    \item A fast constrained sampling-based path planner based on RRT-Connect~\cite{RRTConnect} that satisfies a hard relative transformation constraint between the two end effectors continuously along the entire path.
    \item A constraint-aware state interpolation procedure that guarantees continuous constraint satisfaction between any two valid states, which reuses intermediate IK solutions computed during motion validity checks to densify the planning tree in an efficient manner.
    \item Extensive experimental evaluation on three distinct simulated bimanual platforms, namely KUKA iiwa, Kinova Gen3 and UR5, across multiple environments and transformation constraints, including real-world bimanual Kinova Gen3 demonstrations of tray transport and elongated object manipulation.
\end{enumerate}

\section{RELATED WORK} \label{sec:related-work}
\subsection{Constrained Motion Planning}

Motion planning under geometric constraints requires finding paths on a constraint manifold $\mathcal{X} = \{ q \in \mathcal{Q} \mid F(q) = \mathbf{0} \}$,  a lower-dimensional subset of the ambient C-space~\cite{sampling-based-methods-constraints-survey}, where $q$ stands for the robot configuration while $F(\cdot)$ represents a set of constraints.
Because the constraint manifold has measure zero in~$\mathcal{Q}$, uniformly sampled configurations will almost surely violate the constraint, posing a fundamental challenge to sampling-based planners~\cite{sampling-based-methods-constraints-survey}.

Sampling-based planners~\cite{rrtstar, informed-rrtstar, bitstar} can project randomly drawn samples onto the constraint manifold~\cite{planning-on-constraint-manifold}, or construct local piecewise-linear approximations via numerical continuation~\cite{atlas-state-space-planning}.
Alternatively, constraints can be relaxed to give the feasible set nonzero volume~\cite{bonilla2015}, or enforced directly through trajectory optimization~\cite{traj-opt-manifold}.
While these general techniques are applicable to the bimanual case, they do not exploit the specific kinematic structure of the problem.
IK has long been used to sample constraint-satisfying configurations for dual-arm systems~\cite{gharbi2008, wang2019}, enabling the use of standard sampling-based planning algorithms.
In particular, one can parameterize the constraint manifold by designating one arm as the leader and computing the configuration of the other arm via IK to satisfy the desired relative transformation.
This leader--follower strategy was recently formalized by Cohn et al.~\cite{tedrake-constrained-bimanual}, who leveraged analytic IK solutions to parametrize the constrained C-space and demonstrated its use with sampling-based planners, trajectory optimizers, and the Graph of Convex Sets (GCS) framework. However, their approach relies on the availability of arm-specific analytic IK solutions, and several of pertaining proposed planning methods (IK-PRM, IK-GCS) require offline precomputation in the order of minutes to hours, which become invalidated by any changes in the environment. 
In contrast, our method requires no offline precomputation and achieves planning times significantly faster than comparable methods.

\subsection{Hard Constraints vs. Relaxation}

A central distinction in the constrained planning literature is between methods that enforce hard constraints and those that relax them.
Relaxation-based approaches replace the equality $F(q) = \mathbf{0}$ with a tolerance band $\|F(q)\| < \varepsilon$, giving the feasible set nonzero volume so that standard sampling-based planners can be applied without modification~\cite{sampling-based-methods-constraints-survey}.
Bonilla et al.~\cite{bonilla2015} applied this strategy to bimanual soft-robot manipulation under task constraints, and later extended it to non-interacting constrained control of rigid manipulators~\cite{bonilla2017}.
While relaxation simplifies planning, it shifts the burden of exact constraint satisfaction onto the execution-level controller, and there is no guarantee that the resulting plan can be executed without violating the task constraint.

In contrast, projection-based methods use iterative procedures such as Jacobian pseudo-inverse descent to retract sampled configurations onto the constraint manifold, ensuring that every configuration in the planning tree satisfies $F(q) = \mathbf{0}$ to numerical precision~\cite{planning-on-constraint-manifold, stilman2010}. 
However, each sampled configuration must undergo iterative Jacobian-based retraction steps until it satisfies $F(q)= \mathbf{0}$, requiring repeated evaluation of the constraint Jacobian, a process that is sensitive to the differentiability of $F$ and becomes particularly computationally expensive when analytic Jacobians are unavailable.
Our method enforces hard constraints by construction: the follower arm configuration is always computed via IK to exactly satisfy the relative transformation, so the constraint holds at every state along the path.

\subsection{Closed Kinematic Chains}

Bimanual manipulation of a rigidly grasped object creates a closed kinematic chain: the robot arms and the manipulated object form a kinematic loop whose closure constraint must be maintained throughout the motion~\cite{cortes2005, yakey2001, han2001}. Planning for closed-chain systems is particularly challenging because the loop-closure constraint is an intrinsic geometric constraint that further reduces the effective dimensionality of the C-space. Early work addressed this through active/passive chain decompositions, in which one sub-chain is sampled freely and the remaining sub-chain is closed via IK~\cite{cortes2005, han2001}. Xian et al.~\cite{xian2018} extended closed-chain manipulation planning to multi-arm systems transporting large objects. Our leader--follower parameterization can be viewed as an instance of the active/passive decomposition. Unlike methods that rely on general-purpose numerical projection to close the loop, we directly exploit the manipulator's IK solver, which yields solutions efficiently and allows seeding for continuity between adjacent configurations.

\section{METHODOLOGY} \label{sec:methodology}
\begin{figure*}[t]
  \centering
  \includegraphics[width=\textwidth]{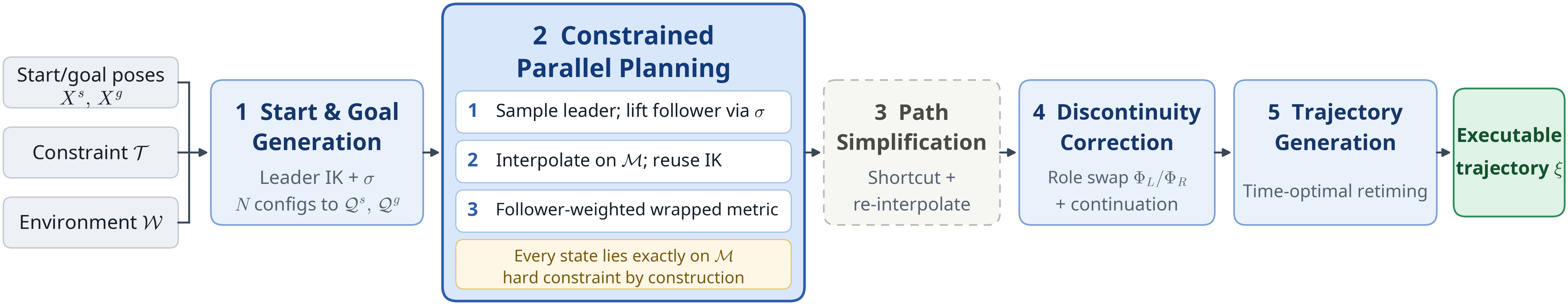}
  \caption{Overview of the proposed constrained bimanual planning pipeline.}
  \label{fig:pipeline}
\end{figure*}

\subsection{Problem Formulation}
\label{subsec:problem-formulation}

Let $\mathcal{X} \subset \mathbb{R}^d$ be the C-space, defined as a compact set of robot configurations $q$.
We denote by $\mathcal{X}_{\mathrm{obs}} \subset \mathcal{X}$ the set of configurations in collision with obstacles or other robot links, and define the collision-free C-space as $\mathcal{X}_{\mathrm{free}} := \mathcal{X} \setminus \mathcal{X}_{\mathrm{obs}}$.
A kinematic constraint $F(q) = 0$ further restricts the set of admissible configurations, inducing a nonlinear submanifold
\begin{equation}
    \mathcal{X}_{\mathrm{valid}} := \{ q \in \mathcal{X}_{\mathrm{free}} \mid F(q) = \textbf{0} \},
    \label{eq:xvalid}
\end{equation}
which captures all configurations that are simultaneously collision-free and satisfy the prescribed kinematic constraint.
Planning then comes down to finding a continuous path $\gamma(t): t\in[0, 1] \rightarrow \mathcal{X}_{\textrm{valid}}$ with $\{\gamma(0) = q^{\mathrm{s}}, \gamma(1) = q^{\mathrm{g}}\}$.

We consider a bimanual robotic system consisting of a \textit{leader} arm and a \textit{follower} arm, each with $n=\{6, 7\}$ degrees of freedom. Let $\mathbf{q}_L \in \mathcal{C}_L$ and $\mathbf{q}_F \in \mathcal{C}_F$ denote the configurations of the leader and follower arms, respectively, and let $\mathbf{q} = (\mathbf{q}_L, \mathbf{q}_F)$ denote the combined system configuration, which lives in the $2n$-dimensional space $\mathcal{C} := \mathcal{C}_L\times\mathcal{C}_F$.
The forward kinematics mappings $\text{FK}_L : \mathcal{C}_L \rightarrow SE(3)$ and $\text{FK}_F : \mathcal{C}_F \rightarrow SE(3)$ yield the end-effector poses of the leader and follower arms in a common world frame.

The desired rigid transformation $\mathcal{T} \in SE(3)$ between the two end effectors is specified and must be satisfied at every point along the trajectory. The hard constraint is given as:
\begin{equation}
    \text{FK}_F(\mathbf{q}_F) = \text{FK}_L(\mathbf{q}_L) \cdot \mathcal{T}, \quad \forall\, t \in [0, 1].
    \label{eq:hard-constraint}
\end{equation}
Given a start $\mathbf{q}^{\mathrm{s}}$ and a goal $\mathbf{q}^{\mathrm{g}}$, the objective is to find a continuous, collision-free path $\gamma(t)$ satisfying \eqref{eq:hard-constraint} at all times, along which an executable trajectory can be constructed.

\subsubsection{The constraint manifold}
Equation~\eqref{eq:hard-constraint} is equivalent to the vanishing of the residual map $F:\mathcal{C}\to\mathbb{R}^6$,
\begin{equation}
    F(\mathbf{q}) \;=\; \log\!\Big(\big(\text{FK}_L(\mathbf{q}_L)\,\mathcal{T}\big)^{-1}\,\text{FK}_F(\mathbf{q}_F)\Big)^{\vee}\in\mathfrak{se}(3),
    \label{eq:residual}
\end{equation}
where $\log(\cdot)^{\vee}$ maps an element of $SE(3)$ to its $6$-vector of exponential coordinates, so that $F(\mathbf{q})=\mathbf{0}$ iff the follower end-effector pose equals the constrained target. The valid set of \eqref{eq:xvalid} is then $\mathcal{X}_{\mathrm{valid}} = \mathcal{X}_{\mathrm{free}}\cap\mathcal{M}$ with
\begin{equation}
    \mathcal{M} := F^{-1}(\mathbf{0}) = \{\mathbf{q}\in\mathcal{C} \mid \text{FK}_F(\mathbf{q}_F) = \text{FK}_L(\mathbf{q}_L)\,\mathcal{T}\}.
    \label{eq:manifold}
\end{equation}
Because $\mathcal{T}$ is constant, differentiating \eqref{eq:hard-constraint} shows that the two end-effector \emph{spatial twists} coincide,
\begin{equation}
    J^{s}_F(\mathbf{q}_F)\,\dot{\mathbf{q}}_F = J^{s}_L(\mathbf{q}_L)\,\dot{\mathbf{q}}_L,
    \label{eq:twist-equality}
\end{equation}
where $J^{s}_L, J^{s}_F$ are the leader and follower spatial Jacobians. Equation~\eqref{eq:twist-equality} identifies the constraint Jacobian as $DF = [\,-J^{s}_L,\; J^{s}_F\,]$, which has full rank $6$ whenever the follower Jacobian $J^{s}_F$ is of full row rank.
By the regular value theorem, $\mathcal{M}$ is then a smooth embedded submanifold of dimension
\begin{equation}
    \dim\mathcal{M} = 2n - 6,
    \label{eq:manifold-dim}
\end{equation}
i.e.\ $8$ for the $7$-DoF arms, and $6$ for the $6$-DoF arms. Since $6\le 2n$, $\mathcal{M}$ is a measure-zero subset of $\mathcal{C}$: uniformly sampled configurations violate the constraint almost surely~\cite{sampling-based-methods-constraints-survey}, which is the central difficulty we address. The rank condition on $J^{s}_F$ also localizes the non-smooth points of $\mathcal{M}$ at follower kinematic singularities, a fact we exploit in Sec.~\ref{subsec:sensitivity}.

\subsection{Constraint-Manifold Parameterization}
\label{subsec:config-parameterization}

Rather than sampling in $\mathcal{C}$ and projecting onto $\mathcal{M}$, we parameterize $\mathcal{M}$ directly. A configuration $\mathbf{q} = (\mathbf{q}_L, \mathbf{q}_F)$ is valid if and only if the follower end-effector pose satisfies the desired transformation constraint with respect to the leader end-effector pose. Given the leader configuration $\mathbf{q}_L$, the required follower end-effector pose $\mathbf{X}_F^* \in SE(3)$ is
\begin{equation}
    \mathbf{X}_F^* = \text{FK}_L(\mathbf{q}_L) \cdot \mathcal{T},
    \label{eq:follower-target}
\end{equation}
and the follower configuration is recovered via IK,
\begin{equation}
    \mathbf{q}_F = \text{IK}_F(\mathbf{X}_F^*,\, \mathbf{q}_F^{\text{seed}}),
    \label{eq:ik}
\end{equation}
where $\mathbf{q}_F^{\text{seed}}$ is a seed configuration provided to bias the IK solver toward solutions close to a reference state.

\subsubsection{Parameterization as a chart of $\mathcal{M}$}
Equations~\eqref{eq:follower-target}--\eqref{eq:ik} define the parameterization map
\begin{equation}
    \sigma(\mathbf{q}_L) = \big(\mathbf{q}_L,\; \text{IK}_F(\text{FK}_L(\mathbf{q}_L)\,\mathcal{T};\, \mathbf{q}_F^{\text{seed}})\big),
    \label{eq:section}
\end{equation}
which lifts a leader configuration to a constraint-satisfying bimanual state. The leader joint angles thereby serve as coordinates on $\mathcal{M}$, so any planner operating in $\mathcal{C}_L$ and lifting through $\sigma$ searches directly on $\mathcal{M}$. In contrast to atlas- and projection-based constrained planners~\cite{atlas-state-space-planning, planning-on-constraint-manifold}, which build local charts by numerical continuation or retract samples with iterative Jacobian steps, the chart here is evaluated by a single call to the manipulator's own IK solver. The construction is agnostic to how $\text{IK}_F$ is realized: an analytic solution where available, an iterative optimization-based solver~\cite{drake, pinocchio, frax} or an IKFast~\cite{IKFast}-generated solver otherwise.

\subsubsection{Role of the seed}
For a fixed leader configuration, the follower IK solution set of \eqref{eq:ik} has dimension $\dim\mathcal{M} - n = n-6$: a finite set of at most eight branches for the $6$-DoF, and a one-dimensional self-motion manifold for the redundant $7$-DoF manipulators. The seed $\mathbf{q}_F^{\text{seed}}$ selects one element of this set, fixing the IK branch and, in the redundant case, the redundancy resolution. Seeding the solver with the previous state makes $\sigma$ vary continuously along a single branch, minimizing discontinuities between adjacent configurations; this is what makes the parameterization usable as the backbone of a sampling-based planner, during both interpolation (Sec.~\ref{sec:interpolation}) and discontinuity correction (Sec.~\ref{subsec:postproc}).

\subsubsection{Constraint satisfaction by construction}
Every state produced by \eqref{eq:section} satisfies $F(\mathbf{q})=\mathbf{0}$ up to the IK solver tolerance, since the follower pose is solved to match $\mathbf{X}_F^*$ exactly. The hard constraint therefore holds at \emph{every} state the planner inserts into the tree, a qualitative departure from methods that enforce the constraint only at sampled configurations and interpolate freely between them, where it may be violated arbitrarily along an edge~\cite{tedrake-constrained-bimanual}.

\subsection{Follower Sensitivity and C-Space Discontinuities}
\label{subsec:sensitivity}

While $\sigma$ places every state exactly on $\mathcal{M}$, it is not equally well-conditioned everywhere. Because the follower configuration is recovered by IK, a small change in the leader configuration can force a disproportionately large change for the follower: near a follower kinematic singularity, it undergoes a large joint-space motion to keep tracking the leader. These large motions are the C-space discontinuities characteristic of constrained bimanual planning; jumps induced by an ill-conditioned parameterization, occurring exactly at singular points of $\mathcal{M}$ identified in Sec.~\ref{subsec:problem-formulation}.

This analysis directly motivates three components of our pipeline. First, the number of interpolation steps between two states is scaled with their distance (Sec.~\ref{sec:interpolation}). Second, samples whose follower configuration lies within a margin of a kinematic singularity are rejected before insertion into the tree. Third, the nearest-neighbor metric used by the planner up-weights the follower joints, so that connections are attempted preferentially between states along which the follower arm moves least. Concretely, we use a wrapped, weighted angular metric:
\begin{equation}
    d(\mathbf{q}, \mathbf{q}') = \big\| \mathbf{w} \odot (\mathbf{q}' \ominus \mathbf{q}) \big\|,
    \quad
    (\mathbf{q}'\ominus\mathbf{q})_i = \operatorname{atan2}\!\big(\sin\delta_i, \cos\delta_i\big),
    \label{eq:metric}
\end{equation}
with $\delta_i = q'_i - q_i$, weights $\mathbf{w}$ that emphasize the follower joints, and $\odot$ the elementwise product. The wrapped difference $\ominus$ respects the circular topology of the continuous joints, so that configurations differing by a full revolution are correctly treated as identical.

\subsection{Start and Goal Configuration Generation}
\label{subsec:start-goal}
Motion planning queries may specify the start and goal either as poses in $SE(3)$ or directly as configurations; since the objective is to move the end effectors between specified poses, planning is posed in C-space regardless of the input representation.
To mitigate the difficulty introduced by the hard constraint, which renders a large fraction of the C-space invalid and makes connectivity between valid states challenging, we generate $N$ distinct configurations for each specified start and goal pose.
Formally, for a start pose $\mathbf{X}^{\mathrm{s}}$ we take the leader preimages $\{\mathbf{q}_L^{\mathrm{s},i}\} = \text{IK}_L(\mathbf{X}^{\mathrm{s}})$, the discrete IK branches for the non-redundant arm, or samples of the redundancy parameter for the redundant arm, and lift each through \eqref{eq:section} to obtain a set of manifold entry points $\mathcal{Q}^{\mathrm{s}} = \{\sigma(\mathbf{q}_L^{\mathrm{s},i})\}$; the goal set $\mathcal{Q}^{\mathrm{g}}$ is generated analogously.
Each of these configurations achieves the same end-effector pose but with different joint values.
This strategy substantially accelerates planning, as it seeds the search from several branches of $\mathcal{M}$ at once, increasing the probability that the start and goal lie in a common connected component of $\mathcal{X}_{\mathrm{valid}}$ and that a connectable pair is found rapidly.

\subsection{State Sampling}
\label{subsec:sampling}
We implement a custom valid-state sampler within the OMPL framework~\cite{ompl}, supporting two sampling modes:

\begin{itemize}
    \item C-space sampling: The leader's configuration is first sampled uniformly on $\mathcal{C}_L$, subject to joint limits. The follower's configuration is then computed via the parameterization of Sec.~\ref{subsec:config-parameterization}. Because the leader coordinates \emph{are} chart coordinates of $\mathcal{M}$, uniform sampling in $\mathcal{C}_L$ directly yields coverage of the manifold.
    \item Cartesian sampling: End-effector poses are sampled uniformly from a bounding volume enclosing the reachable workspace in $SE(3)$, and subsequently mapped to C-space via IK. This mode additionally supports the specification of an absolute orientation constraint in the world frame, ensuring that all sampled states satisfy a desired global end-effector orientation.
\end{itemize}

To further avoid wasted computation, a leader configuration whose end-effector position falls outside a precomputed bounding box of the bimanually reachable workspace is rejected before the follower IK is attempted, since a valid follower parameterization is extremely unlikely to exist there.

\subsection{Constraint-Aware State Interpolation}
\label{sec:interpolation}

A key contribution of our method is constraint-aware interpolation that guarantees satisfaction of the hard transformation constraint~\eqref{eq:hard-constraint} at every point along the interpolated path.
Naive linear or spline interpolation in the C-space does not preserve the transformation constraint between states and is thus unsuitable.
We instead trace a curve on $\mathcal{M}$ by a predictor--corrector scheme that is a specialization of numerical continuation~\cite{atlas-state-space-planning} to the chart $\sigma$. Given two valid states $\mathbf{q}^a$ and $\mathbf{q}^b$, the distance $l = d(\mathbf{q}^a, \mathbf{q}^b)$ is first computed and the interval is discretized into $K$ steps, with $K$ chosen proportional to $l$ so that individual steps remain small where the parameterization is stiff. At each step $k$, the \emph{predictor} advances the leader along a linear interpolation,
\begin{equation}
    \mathbf{q}_L^{(k)} = (1 - \alpha_k)\,\mathbf{q}_L^a + \alpha_k\,\mathbf{q}_L^b, \quad \alpha_k = \frac{k}{K},
    \label{eq:predictor}
\end{equation}
and the \emph{corrector} recovers the follower configuration $\mathbf{q}_F^{(k)}$ via the parameterization~\eqref{eq:follower-target}--\eqref{eq:ik}, warm-started from the previous follower state $\mathbf{q}_F^{(k-1)}$ to remain on the same branch. Unlike tangent-space continuation, which requires an orthogonal retraction per step, each corrector here is a single IK evaluation and returns a state that lies \emph{exactly} on $\mathcal{M}$ by construction. The resulting sequence $\{(\mathbf{q}_L^{(k)}, \mathbf{q}_F^{(k)})\}_{k=0}^{K}$ constitutes a constraint-satisfying discretized path segment.

IK evaluations during interpolation are expensive, so all intermediate states generated during a motion validity check are retained and inserted into the planning tree.
This densifies the tree, which both accelerates the discovery of a complete path and reduces the configuration-space discontinuities between adjacent states in the final path.
In effect, motion validation doubles as informed sampling, populating $\mathcal{M}$ along already-explored, constraint-satisfying directions at no additional IK cost.
Interpolation can additionally be performed in Cartesian space, enabling the simultaneous enforcement of both the relative transformation constraint and a desired absolute end-effector orientation.

\subsection{Path Planning Algorithm}
\label{subsec:planning}

Path planning is performed directly in the bimanual C-space, where every state in the tree satisfies the transformation constraint~\eqref{eq:hard-constraint} by construction.
We employ RRT-Connect~\cite{RRTConnect} as the underlying planner, implemented within OMPL~\cite{ompl}, with the wrapped, follower-weighted metric~\eqref{eq:metric} governing nearest-neighbor queries and the constraint-aware interpolation of Sec.~\ref{sec:interpolation} governing edge validation.
While any sampling-based method can be used for path planning with our pipeline, our evaluations have shown that RRT-Connect is the most efficient for the considered problem, among BIT*~\cite{bitstar}, RRT*~\cite{rrtstar}, Informed RRT*~\cite{informed-rrtstar}, BiEST~\cite{biest}, SBL~\cite{sbl}, STRIDE~\cite{stride}, and RRT\#~\cite{rrtsharp}.

\subsection{Path Post-Processing and Trajectory Generation}
\label{subsec:postproc}

The complete planning pipeline is summarized in Fig~\ref{fig:pipeline}.
After a raw path is found, it undergoes the following post-processing stages before a trajectory is returned.

\subsubsection{Path Simplification}
The raw path is shortcut by iteratively attempting to replace sub-sequences of states with direct constraint-satisfying connections, followed by re-interpolation between the remaining waypoints using the procedure of Sec.~\ref{sec:interpolation}.

\subsubsection{Discontinuity Detection and Correction}
Due to the hard constraint, IK solutions for adjacent states may occasionally exhibit discontinuities in the C-space, which must be removed before a smooth trajectory can be constructed. We exploit the fact that the role assignments induce two parameterizations of the same manifold, $\Phi_L$ (leader $= L$) and $\Phi_R$ (leader $= R$), which together cover $\mathcal{M}$. The correction procedure first attempts to resolve a discontinuity by treating each arm alternately as the leader, and selecting the solution closest to the preceding state. If the discontinuity cannot be resolved in this manner due to its magnitude, additional constraint-satisfying states are inserted between the problematic pair via the continuation procedure of Sec.~\ref{sec:interpolation}.

\subsubsection{Trajectory Generation}
A time-parameterized trajectory is generated from the processed path using TOPP-RA~\cite{toppra}, a time-optimal path parameterization subject to joint velocity and acceleration constraints. If unresolved discontinuities remain in the path, TOPP-RA will produce a trajectory that violates the transformation constraint; such trajectories are detected by evaluating the residual~\eqref{eq:residual} along the retimed trajectory and are rejected. Only trajectories that pass this final constraint-satisfaction check are returned.

\section{EXPERIMENTAL RESULTS} \label{sec:exp-results}
We demonstrate our constrained bimanual motion planning pipeline on three distinct bimanual robotic platforms: (1) two KUKA iiwa 7-DoF arms, (2) two Kinova Gen3 7-DoF arms, and (3) two UR5 6-DoF arms. Experiments with the KUKA iiwa configuration replicate the setup of~\cite{tedrake-constrained-bimanual} to enable direct performance comparison. Experiments with the UR5 were conducted in simulation while experiments with Kinova Gen3 were conducted in simulation as well as on physical hardware. Simulation trials span three distinct environments and three different end-effector transformation constraints, while real-world trials evaluate performance under two transformation constraints. All experiments were conducted on the Intel Core i7-11800H CPU.

\subsection{Comparison Against Prior Work on the KUKA iiwa}

\begin{figure}[t]
    \centering
    \begin{subfigure}[b]{0.33\columnwidth}
        \includegraphics[width=\linewidth]{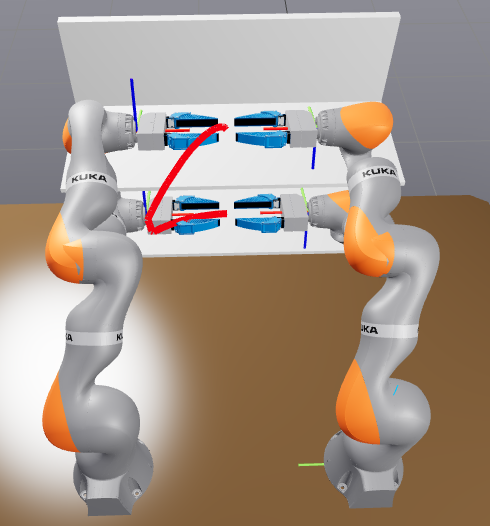}
        \label{fig:ttm-kuka}
    \end{subfigure}%
    \hfill%
    \begin{subfigure}[b]{0.31\columnwidth}
        \includegraphics[width=\linewidth]{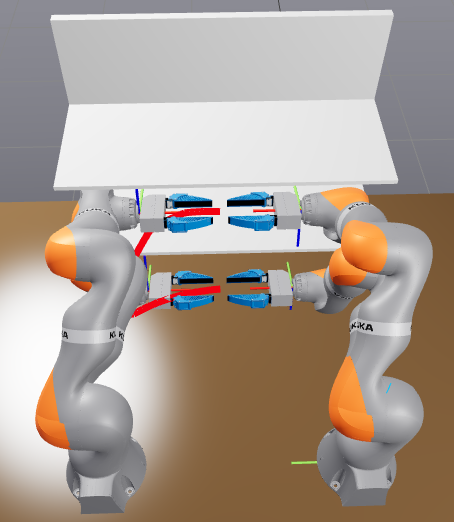}
        \label{fig:mtb-kuka}
    \end{subfigure}%
    \hfill%
    \begin{subfigure}[b]{0.35\columnwidth}
        \includegraphics[width=\linewidth]{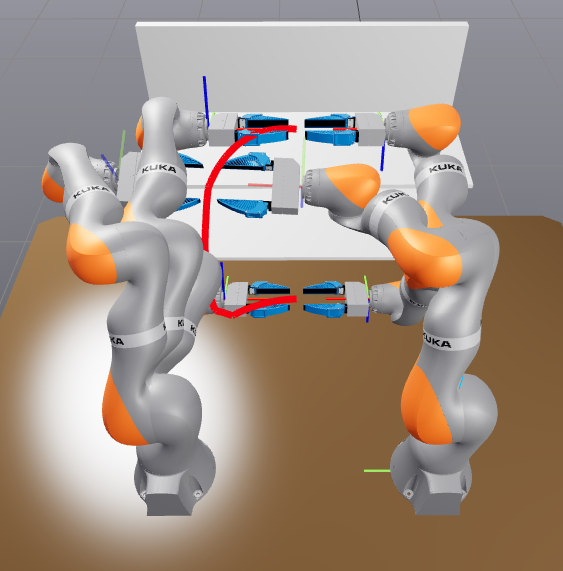}
        \label{fig:btt-kuka}
    \end{subfigure}
    \caption{Trajectories for three different start-goal configurations using the KUKA iiwa bimanual setup.
    }
    \label{fig:kuka-exp}
\end{figure}

To assess the performance of our method against the state of the art, we evaluate on the benchmark introduced in~\cite{tedrake-constrained-bimanual}, which, to our knowledge, represents the only directly comparable approach to constrained bimanual motion planning that enforces the relative end-effector transformation as a hard constraint throughout the entire trajectory. In this benchmark, two arms must collaboratively transport an object among a set of shelves while avoiding collisions, subject to a fixed rigid transformation constraint between end effectors. Trials are conducted for three configurations: \textit{Top to Middle}, \textit{Middle to Bottom}, and \textit{Bottom to Top}. The computed trajectories for three different configurations are displayed in Fig. \ref{fig:kuka-exp}. We use the analytic IK solution for the KUKA iiwa provided in~\cite{tedrake-constrained-bimanual} to ensure a fair and consistent comparison.

We compare against IK-BiRRT~\cite{tedrake-constrained-bimanual}, as both methods employ a bidirectional RRT as the underlying sampling-based planner, making this the most direct comparison. For completeness, we also report the results of therein presented IK-GCS and IK-PRM methods, though we note that these methods require substantial offline precomputation: IK-GCS requires approximately 5.1 hours to construct convex collision-free regions, and IK-PRM requires over 44 minutes to construct the roadmap. Crucially, any change in the environment, such as the introduction of a new obstacle, invalidates this precomputation entirely, necessitating a full rebuild. Our method requires no offline precomputation and can plan immediately in unseen environments.

The results for IK-BiRRT, IK-PRM, and IK-GCS are reported as published in~\cite{tedrake-constrained-bimanual}, since their implementation is not publicly available.
Average path lengths and online planning times across 10 runs per configuration are reported in Table~\ref{tab:combined_results}. Our method achieves planning times that are $19.4$ times faster than IK-BiRRT on average, and produces shorter paths in $2$ out of $3$ configurations.
Furthermore, the parameterization method in~\cite{tedrake-constrained-bimanual} exhibited constraint violations of up to $1 \times 10^{-3}$~cm, whereas our parameterization maintained the violation below $1.7 \times 10^{-17}$~cm, satisfying the rigid transformation hard constraint up to numerical precision.

\begin{table}[!t]
\centering
\caption{Path length (in C-space) and online planning time (in seconds) on KUKA iiwa benchmark. Methods above the dashed line require offline computation; methods below are fully online. \textbf{Bold} marks the best value in each column; \underline{underline} marks the best among the fully-online methods.}
\label{tab:combined_results}
\begin{tblr}{
  width = \linewidth,
  colspec = {llX[c]X[c]X[c]},
  cell{2}{1} = {r=2}{valign=m}, cell{4}{1} = {r=2}{valign=m},
  cell{6}{1} = {r=2}{valign=m}, cell{8}{1} = {r=2}{valign=m},
  hline{1,2,10} = {solid},
  hline{6} = {dashed},
  row{1} = {abovesep=1pt, belowsep=1pt},
  row{5} = {belowsep=3pt},
  row{6} = {abovesep=3pt},
}
\textbf{Method} & \textbf{Metric} & \textbf{T$\to$M} & \textbf{M$\to$B} & \textbf{B$\to$T} \\
IK-PRM~\cite{tedrake-constrained-bimanual}   & Length   & 4.67 & 8.93 & 9.21 \\
                                              & Time (s) & \textbf{0.46} & 0.64 & \textbf{0.61} \\
IK-GCS~\cite{tedrake-constrained-bimanual}   & Length   & \textbf{2.09} & \textbf{3.32} & \textbf{5.62} \\
                                              & Time (s) & 3.41 & 2.32 & 3.32 \\
IK-BiRRT~\cite{tedrake-constrained-bimanual} & Length   & 9.91 & 8.69 & \underline{11.42} \\
                                              & Time (s) & 49.42 & 52.53 & 54.10 \\
Ours                                          & Length   & \underline{8.70} & \underline{6.55} & 20.57 \\
                                              & Time (s) & \underline{2.66} & \textbf{0.45} & \underline{4.948} \\
\end{tblr}
\end{table}

\subsection{Kinova Gen3 and UR5 Simulation Experiments}

To further evaluate the performance of our method, we conduct a comprehensive set of simulation experiments using two Kinova Gen3 7-DoF and two UR5 6-DoF arms.

\subsubsection{Experimental Setup}

We consider three simulation environments of increasing obstacle density.
The first environment contains no collision obstacles, serving as a baseline to assess pure kinematic feasibility. The second environment contains sparsely placed collision spheres, providing moderate obstacle density. The third environment contains densely placed collision spheres, with minimal clearance between them, representing a highly cluttered workspace.

Three distinct rigid transformation constraints between the leader and follower end effectors are evaluated, selected to reflect practically relevant bimanual manipulation scenarios:
\begin{itemize}
    \item \textbf{$\mathcal{T}_1$}: The follower arm is maintained parallel to the leader arm. This is well-suited for non-prehensile manipulation tasks, such as collaboratively lifting and transporting a box without graspable handles, analogous to how a human would carry a large object with both hands.
    \item \textbf{$\mathcal{T}_2$}: The follower arm is oriented such that its end effector faces toward the leader arm. This constraint is applicable to collaborative pick-and-place operations, as demonstrated in the experiments shown in Fig.~\ref{fig:real-world-tacna}.
    \item \textbf{$\mathcal{T}_3$}: Similar to $\mathcal{T}_1$, but with the position offset defined along a different axis. This constraint is particularly suited for transporting elongated objects that exceed the reach of a single arm, as demonstrated in Fig.~\ref{fig:real-world-stap}.
\end{itemize}

Start and goal configurations were sampled from the reachable bimanual workspace. We do not verify path existence for each start–goal pair prior to planning, as doing so would be computationally prohibitive. To ensure non-trivial planning queries, start and goal states are paired such that a minimum separation is enforced in both C-space and Cartesian space, thereby excluding trivially close configurations that would not meaningfully challenge the planner.

To the best of our knowledge, no general-purpose, off-the-shelf constrained bimanual motion planner is publicly available for comparison on this problem. The IK-BiRRT method of~\cite{tedrake-constrained-bimanual} is not open-sourced, precluding its direct application to our experimental setup. We therefore implement a naive baseline for comparison, which plans a collision-free path in the leader's arm C-space without accounting for the transformation constraint. In a post-processing step, for each state along the found path, the follower arm configuration is computed via IK to satisfy the hard transformation constraint.

\subsubsection{Results}

\begin{table*}[t]
\centering
\caption{%
  Full quantitative results on both platforms. \emph{PSR}: pipeline success rate
  (\%). $t_\text{plan}$: average path-planning time (s) over successful paths, $\ell$: average path length in the full configuration space.
  $t_\text{pipe}$: average end-to-end pipeline time (s) for successful trajectories.
  \textbf{Bold} PSR marks our method. $\dagger$~no valid trajectory for any query;
  ``---'' no successful pipeline.}
\label{tab:main-compact}
\setlength{\tabcolsep}{3pt}
\renewcommand{\arraystretch}{1.05}
{\footnotesize
\begin{tabular*}{\textwidth}{@{\extracolsep{\fill}}
  l l
  *{3}{S[table-format=3.1] S[table-format=2.2] S[table-format=2.1] S[table-format=2.2]}
  @{}}
\toprule
 & & \multicolumn{4}{c}{No obst.}
     & \multicolumn{4}{c}{Sparse}
     & \multicolumn{4}{c}{Dense} \\
\cmidrule(lr){3-6}\cmidrule(lr){7-10}\cmidrule(lr){11-14}
Constr. & Planner
  & {PSR} & {$t_\text{plan}$} & {$\ell$} & {$t_\text{pipe}$}
  & {PSR} & {$t_\text{plan}$} & {$\ell$} & {$t_\text{pipe}$}
  & {PSR} & {$t_\text{plan}$} & {$\ell$} & {$t_\text{pipe}$} \\
\midrule
\multicolumn{14}{@{}l}{\textbf{Kinova Gen3}} \\
\multirow{2}{*}{$\mathcal{T}_1$}
  & Ours     & \bfseries 98.0 & 0.98 & 10.3 & 3.75 & \bfseries 80.0 & 22.02 & 26.9 & 30.01 & \bfseries 86.7 & 14.58 & 29.1 & 24.98 \\
  & Baseline & 32.0 & 1.78 & 16.8 & 3.57 & 0.0\rlap{$^\dagger$} & 3.07 & 17.8 & {---} & 0.0\rlap{$^\dagger$} & 2.97 & 22.8 & {---} \\
\addlinespace[1.5pt]
\multirow{2}{*}{$\mathcal{T}_2$}
  & Ours     & \bfseries 93.0 & 1.06 & 9.5 & 2.92 & \bfseries 70.0 & 29.94 & 31.2 & 27.85 & \bfseries 93.3 & 17.63 & 15.5 & 21.10 \\
  & Baseline & 49.0 & 2.47 & 14.1 & 3.18 & 6.7 & 3.62 & 20.9 & 4.97 & 23.3 & 3.54 & 18.4 & 3.55 \\
\addlinespace[1.5pt]
\multirow{2}{*}{$\mathcal{T}_3$}
  & Ours     & \bfseries 98.0 & 0.93 & 7.0 & 3.33 & \bfseries 86.7 & 6.61 & 14.6 & 12.12 & \bfseries 100.0 & 2.70 & 12.6 & 6.45 \\
  & Baseline & 58.0 & 2.34 & 13.1 & 3.74 & 3.3 & 3.78 & 19.8 & 4.04 & 16.7 & 3.28 & 19.4 & 4.47 \\
\midrule
\multicolumn{14}{@{}l}{\textbf{UR5}} \\
\multirow{2}{*}{$\mathcal{T}_1$}
  & Ours     & \bfseries 99.0 & 4.64 & 6.3 & 19.40 & \bfseries 60.0 & 25.16 & 12.0 & 52.20 & \bfseries 80.0 & 24.15 & 18.1 & 52.18 \\
  & Baseline & 15.0 & 0.94 & 20.3 & 8.14 & 0.0\rlap{$^\dagger$} & {---} & {---} & {---} & 6.7 & 1.01 & 17.0 & 2.42 \\
\addlinespace[1.5pt]
\multirow{2}{*}{$\mathcal{T}_2$}
  & Ours     & \bfseries 92.0 & 1.70 & 5.4 & 17.87 & \bfseries 16.7 & 67.83 & 32.8 & 76.47 & \bfseries 86.7 & 21.71 & 12.6 & 45.75 \\
  & Baseline & 7.0 & 0.94 & 20.8 & 4.26 & 0.0\rlap{$^\dagger$} & 1.04 & 20.1 & {---} & 0.0\rlap{$^\dagger$} & 0.99 & 16.5 & {---} \\
\addlinespace[1.5pt]
\multirow{2}{*}{$\mathcal{T}_3$}
  & Ours     & \bfseries 99.0 & 1.09 & 2.8 & 5.97 & \bfseries 100.0 & 1.85 & 4.4 & 12.81 & \bfseries 100.0 & 1.42 & 3.1 & 7.80 \\
  & Baseline & 15.0 & 0.94 & 12.2 & 4.29 & 0.0\rlap{$^\dagger$} & 1.02 & 14.6 & {---} & 3.3 & 0.98 & 15.6 & 2.93 \\
\bottomrule
\end{tabular*}%
}
\end{table*}

Comprehensive analyses of experimental results for Kinova Gen3 and UR5 platforms are presented in Table~\ref{tab:main-compact}.
We report the pipeline success rate (PSR), defined as the fraction of start--goal queries for which the full planning pipeline produces a valid, executable trajectory. Experiments were conducted with 100 randomly sampled queries in the obstacle-free environment and 30 queries in each of the obstacle environments, with a maximum allowed planning time of 120\,s per query.

The results reveal a substantial performance gap between our method and the naive baseline across all environments and transformation constraints.
When the follower configuration is subsequently computed via IK in the post-processing step to satisfy the relative transformation constraint, two failure modes arise.
First, the IK solver frequently produces discontinuous solutions between adjacent states along the path, resulting in large C-space jumps that cannot be resolved into a smooth, executable trajectory.
Second, a follower configuration that satisfies the transformation constraint may place the follower arm in collision with obstacles or with the leader arm itself, rendering the entire path invalid.
In contrast, our method plans directly in the full bimanual C-space, where every sampled state is guaranteed to be collision-free and correctly parameterized under the transformation constraint.
This ensures that the planner only expands states for which a valid follower configuration exists, substantially increasing the likelihood of finding a path that is both constraint-satisfying and collision-free throughout.

This fundamental difference is reflected in the results.
In the obstacle-free environment, the baseline achieves moderate PSR, since collisions between the follower arm and the environment are not a concern.
However, as obstacle density increases, the baseline performance degrades sharply.
In the obstacle environments, the baseline frequently produces zero successful trajectories across all transformation constraints.
Our method maintains substantially higher PSR across all tested configurations. We additionally ablated the path-simplification stage of the pipeline. Since simplification is not required for constraint satisfaction, omitting it reduces the average pipeline time by $20\%$, at the cost of paths that are on average $32\%$ longer in the C-space, making it an optional stage that can be disabled in time-critical applications.

\subsection{Real-World Experiments on the Kinova Gen3 Platform}

\begin{figure}[!t]
    \centering
    \includegraphics[width=1\linewidth]{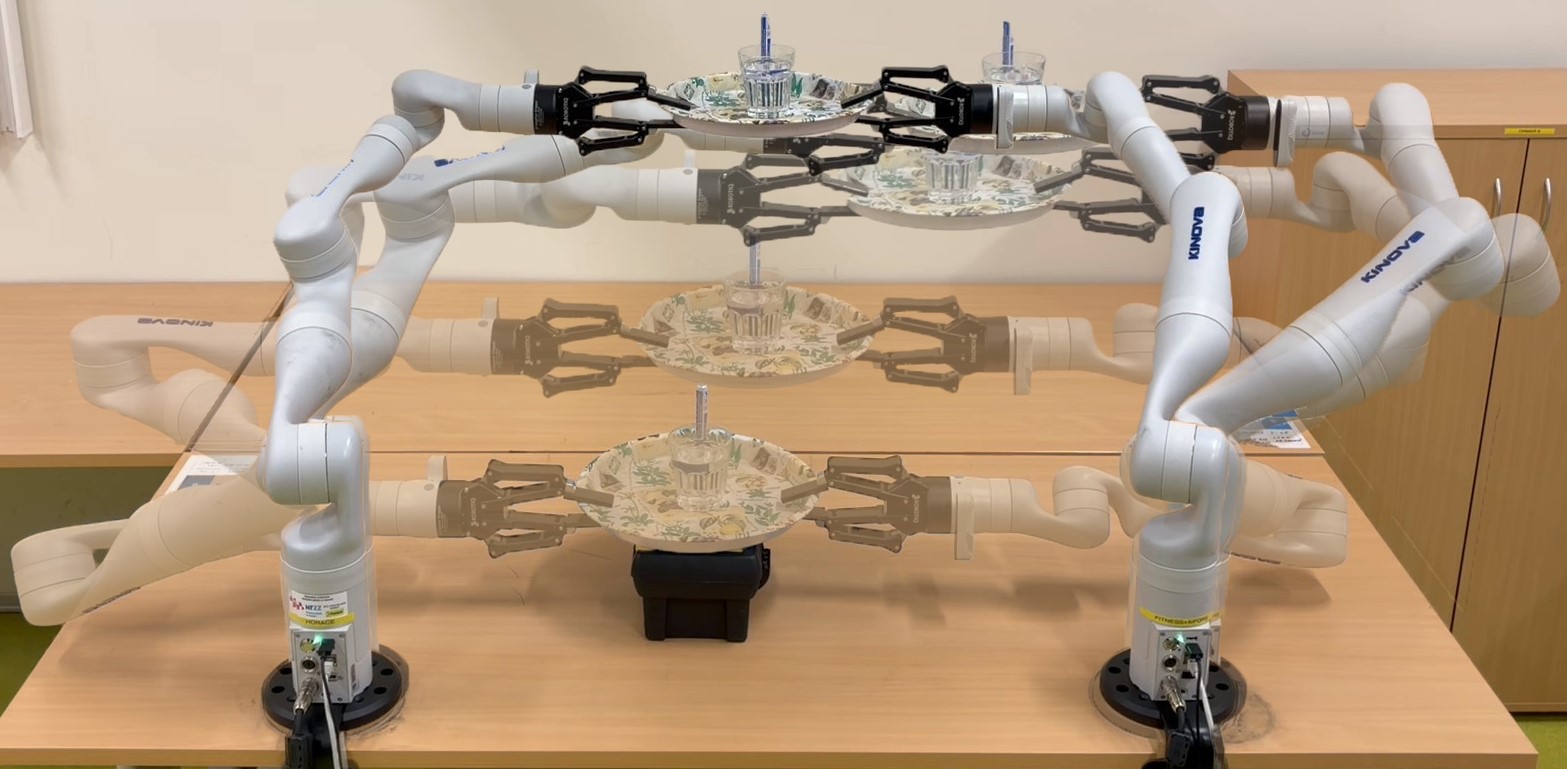}
    \caption{Real-world tray transport experiment. In addition to the relative transform constraint, global orientation constraint needed to be satisfied. The motion is best viewed from supplementary videos.}
    \label{fig:real-world-tacna}
\end{figure}

To validate the practical applicability of our approach, we conducted real-world experiments using two Kinova Gen3 7-DoF manipulators in a laboratory setting. Two distinct transformation constraints are evaluated ($\mathcal{T}_2$ and $\mathcal{T}_3$).

\subsubsection{Tray Transport ($\mathcal{T}_2$)} \label{subsec:tray-transport}

In the first experiment, the follower end effector is oriented to face toward the leader end effector. The task requires the two arms to collaboratively pick up a tray carrying a glass filled with water and an upright marker pen, and transport it to a specified goal pose. The relative transformation must be maintained precisely throughout execution, and the trajectory must be sufficiently smooth to prevent the water from spilling or the marker from toppling. The planned trajectory is visualized in Fig.~\ref{fig:real-world-tacna}. Experiments demonstrated successful task completion, with the tray contents remaining undisturbed throughout the motion, validating constraint satisfaction and smoothness properties.

\subsubsection{Elongated Object Transport ($\mathcal{T}_3$)}

\begin{figure}[!t]
    \centering
    \includegraphics[width=0.8\linewidth]{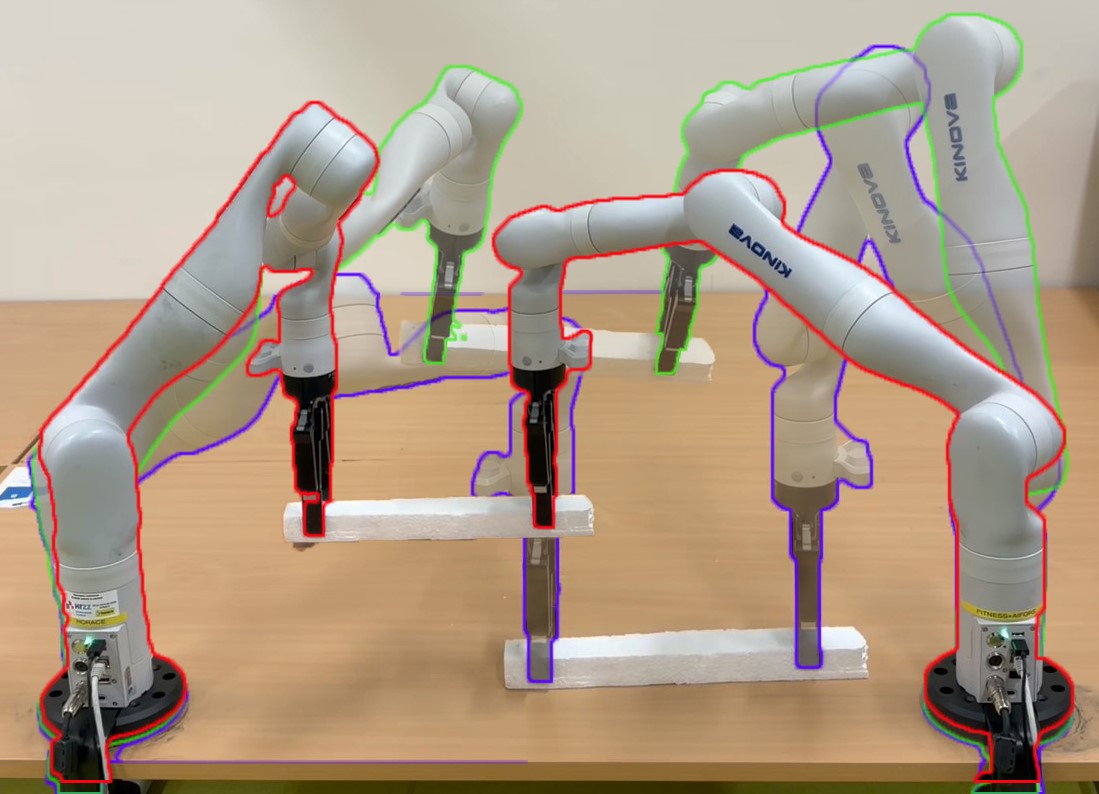}
    \caption{Real-world elongated object transport experiment. Movement sequence: blue $\rightarrow$ green $\rightarrow$ red. The motion is best viewed from supplementary videos.}
    \label{fig:real-world-stap}
\end{figure}

In the second experiment the two end effectors share the same orientation but are offset along a specified axis, appropriate for the collaborative transport of elongated objects which are unfeasible to be carried by single arm because of its dimensions or weight. The task requires the two arms to grasp the object at its initial pose and carry it to a designated goal pose. The planned trajectory is visualized in Fig.~\ref{fig:real-world-stap}. Experiments confirmed successful execution, demonstrating that our planner generalizes to object geometries and constraint configurations that arise naturally in warehouse and industrial manipulation scenarios.

\subsection{Discussion, limitations and future work}
Real-world experiments confirm that planned trajectories are directly executable on physical hardware without post-hoc correction. The hard constraint is maintained continuously along the entire executed trajectory, and the smoothness of the planned paths prevents undesirable dynamic effects during execution, validating that the proposed pipeline transfers successfully from simulation to the real-world.

In all experiments, the bimanual platforms are symmetric, both arms are identical and mounted in mirrored configurations, so the choice of leader and follower does not affect the properties of the method.
We consistently designate the right arm as the leader throughout our experiments. In asymmetric settings, the choice of leader can impact planning efficiency: when one arm operates in a more cluttered region of the workspace, it should be designated as the leader, since the leader arm is sampled freely in C-space and collision checks are cheaper than solving the IK.

The primary computational bottleneck is the numerical IK solver invoked during constraint-aware interpolation. While numerical IK is necessary to ensure generality across manipulators that lack closed-form solutions, it is significantly more expensive than analytic alternatives. An alternative approach, randomly sampling configurations and projecting them onto the constraint manifold via Jacobian-based methods~\cite{sampling-based-methods-constraints-survey} avoids the dependence on IK entirely, but in our evaluation this approach lead to substantially longer planning times due to the cost and potential failure of the projection step, as well as lowering space exploration due to projection.

A key direction for future work is improving the robustness and efficiency of C-space discontinuity resolution. The current post-processing procedure resolves discontinuities after planning, which can fail when the distance between configurations is large. Finding a strategy that suppresses large C-space discontinuities during planning without sacrificing tree density and planning speed remains an open problem. A solution to this challenge would directly translate into higher success rates, particularly in cluttered environments.

\section{CONCLUSION} \label{sec:conclusion}
We presented a constrained bimanual motion planning pipeline that enforces a hard rigid-body transformation constraint between two end effectors continuously along the entire trajectory. Our method adopts a leader--follower parameterization in which the follower configuration is solved at every state via IK, guaranteeing constraint satisfaction by construction. A constraint-aware interpolation scheme maintains the constraint between any two valid states, and intermediate IK solutions from motion validity checks are reused to accelerate the search. The proposed pipeline requires no offline precomputation, is robot-agnostic, and deploys immediately in unseen environments.
The proposed method proved effective on three bimanual platforms.
On the KUKA iiwa benchmark it planned 19.4× faster than the comparable prior method~\cite{tedrake-constrained-bimanual} while guaranteeing continuous constraint satisfaction.
On the Kinova Gen3 and UR5 platforms, simulations showed substantially higher success rates than a naive baseline.
Real-world transport experiments validated direct transfer of planned trajectories to hardware.

\balance

\bibliographystyle{IEEEtran}
\bibliography{literature}

\end{document}